\documentclass{article}

\usepackage{arxiv}
\usepackage[utf8]{inputenc}
\usepackage[T1]{fontenc}
\usepackage{hyperref}
\usepackage{url}
\usepackage{booktabs}
\usepackage{amsmath,amssymb,amsfonts}
\usepackage{microtype}
\usepackage{graphicx}
\usepackage{xcolor}
\usepackage{enumitem}
\usepackage{natbib}
\usepackage{doi}
\let\cite\citep
\graphicspath{{media/}}
\hypersetup{
  colorlinks=true,
  allcolors=black,
  pdftitle={When Better Traffic Forecasts Fail to Improve Signal Control: A Layered Diagnostic Study of Forecast-to-Decision Value},
  pdfauthor={Jianing Long, Xiaobin Li, Wuming Lei, and Weiguang Wang},
  pdfkeywords={traffic forecasting, traffic signal control, decision value, conformal prediction, uncertainty quantification}
}

\renewcommand{\shorttitle}{Forecast-to-Decision Value in Traffic Control}
\renewcommand{\headeright}{A Preprint}
\renewcommand{\undertitle}{A Preprint}
\date{October 2, 2026}

\title{When Better Traffic Forecasts Fail to Improve Signal Control:\\
A Layered Diagnostic Study of Forecast-to-Decision Value}

\author{
  Jianing Long\textsuperscript{*} \\
  School of Transportation Engineering \\
  East China Jiaotong University \\
  \href{mailto:ljnvs@foxmail.com}{\textcolor{black}{\texttt{ljnvs@foxmail.com}}}
  \And
  Xiaobin Li \\
  School of Transportation Engineering \\
  East China Jiaotong University
  \AND
  Wuming Lei \\
  School of Transportation Engineering \\
  East China Jiaotong University
  \And
  Weiguang Wang \\
  School of Transportation Engineering \\
  East China Jiaotong University
}

\begin{document}
\maketitle

\begin{abstract}
Improved traffic forecasts do not necessarily yield better signal-control decisions. We investigate this gap through a layered diagnostic study using 29 days of reconstructed demand from Xuancheng, China, with seven dates reserved for testing. The framework evaluates point forecasts, conformal intervals, dependence-aware scenarios, and matched closed-loop controllers. Entry-level and movement-level forecasts reduce mean absolute error by 4.03\% and 3.92\%, respectively, relative to historical means. A nominal 90\% conformal interval achieves 90.72\% marginal coverage but only 75.66\% on an ex-post high-demand subset. Interface audits identify decision-time leakage and reveal that only two of nine controlled intersections offer multiple effective actions. We correct the temporal interface and compare causal forecasts with a five-second event oracle using exhaustive joint-action search. A synthetic positive control demonstrates that future information can reduce the internal rollout cost by 61.5\%. On the frozen test dates, however, causal forecasts and the event oracle increase queue vehicle-seconds by 6.09\% and 3.39\% relative to the matched no-future rollout, while the oracle reduces spillback exposure by 3.78\%; paired-day bootstrap intervals cross zero. These findings indicate that forecast value depends on temporal observability, action identifiability, dynamics consistency, and objective alignment. The proposed protocol provides a practical way to diagnose where predictive improvements fail to translate into operational benefits.
\end{abstract}


\section{Introduction}

Short-term traffic forecasting has progressed from historical averages and conventional time-series models to graph neural networks, recurrent architectures, and attention-based multi-horizon systems \cite{vlahogianni2014short,li2018dcrnn,yu2018stgcn,wu2019graphwavenet,lim2021tft}. In parallel, traffic signal control has advanced from fixed plans and pressure-based feedback to model predictive control and multi-agent reinforcement learning \cite{varaiya2013maxpressure,wongpiromsarn2012distributed,wei2019presslight,wei2019colight,chen2020thousand}. These two research streams invite an intuitive assumption: if future demand is predicted more accurately, a controller should make better decisions and produce less congestion.

The assumption is not generally valid. Forecasting and control optimize different losses. Mean absolute error (MAE) weights errors across locations, horizons, and demand levels, whereas traffic-control losses concentrate near bottlenecks, storage limits, competing phases, and switching constraints. A forecast may improve primarily where no feasible action changes. Conversely, a numerically small error near a critical queue threshold can reverse a signal decision. Marginal calibration and joint scenario quality add further distinctions: a well-calibrated interval for each target does not guarantee a realistic multi-location demand trajectory, and better joint scenarios are useful only if the controller has a structural mechanism for exploiting them.

This mismatch resembles the broader predict-then-optimize problem \cite{wilder2019decision,elmachtoub2022smart}, but signal control introduces specific failure modes. First, the spatial semantics of the target must match the action space. Total demand at an entry road does not identify whether approaching vehicles will go straight or turn, even though these movements may require different phases. Second, a forecast must be observable at the decision time. A model that uses the completed count for minute $m$ cannot be called at the beginning or middle of minute $m$. Third, the controller must possess positive oracle value: if perfect future information does not improve the same control structure, improving the forecast alone has no credible path to benefit. Fourth, queue, throughput, and spillback objectives may disagree, so ``better control'' is not a scalar concept unless the primary metric is declared in advance.

Most traffic-forecasting papers stop at offline predictive scores, while many control papers assume demand or state information is available inside a simulator. What is missing is an auditable chain linking forecast accuracy, reliability, information availability, action sensitivity, oracle value, and final closed-loop effects under one frozen test protocol. Without this chain, a failed controller can be blamed on the forecast, or a forecast can be credited for a gain caused by a changed controller.

We address this gap with a layered diagnostic study based on the Xuancheng city-scale traffic dataset \cite{ma2026city}. Real traffic records are reconstructed as exogenous source-departure demand for a nine-intersection signalized subnetwork. We evaluate entry-level and movement-level forecasts, conformalized quantile intervals, and dependence-aware scenarios. We then follow a sequence of control experiments that progressively repair the deterministic baseline, spatial target, rollout dynamics, temporal interface, and action search. The final experiment uses the same 60-second rollout horizon and exact effective-action enumeration for a no-future controller, a causal forecast controller, and a five-second event-oracle controller.

The study asks four questions:
\begin{enumerate}[leftmargin=1.6em]
  \item Do offline predictive gains persist in high-demand and joint-risk states?
  \item Are the forecast targets spatially and temporally aligned with signal decisions?
  \item After correcting the interface and action search, does the controller exhibit positive oracle information value?
  \item Why can predictive gains and traffic-control gains differ across dates, metrics, and experimental stages?
\end{enumerate}

\begin{figure}[t]
  \centering
  \includegraphics[width=\textwidth]{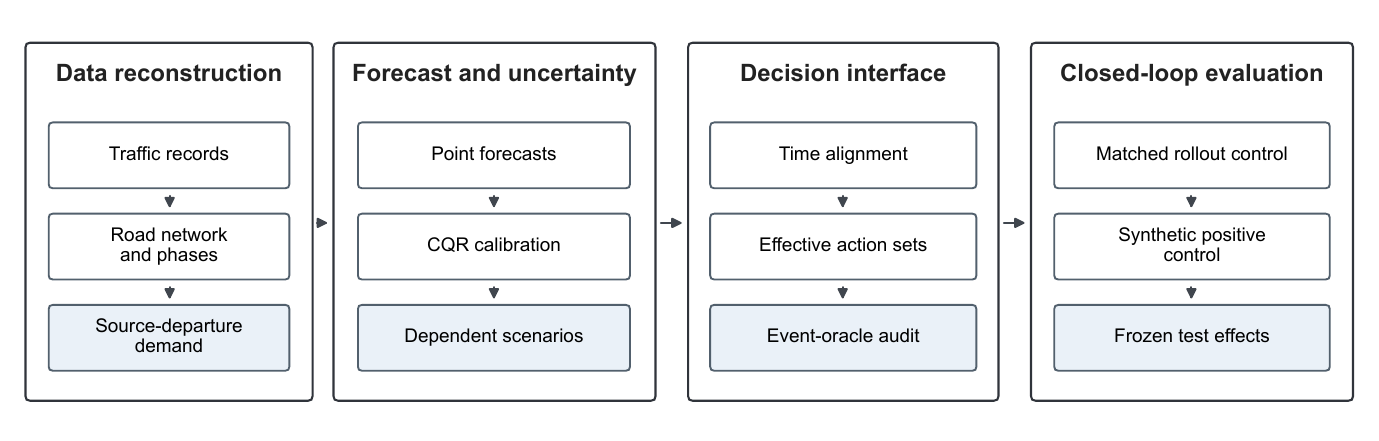}
  \caption{Study workflow across data reconstruction, forecasting and uncertainty estimation, decision-interface auditing, and closed-loop evaluation. Model selection and control tuning use validation data before the frozen test evaluation.}
  \label{fig:framework}
\end{figure}

Our contributions are as follows.
\begin{itemize}[leftmargin=1.4em]
  \item We propose a layered admission protocol spanning forecast quality, temporal observability, action identifiability, oracle structural value, and frozen closed-loop value (Fig.~\ref{fig:framework}).
  \item We provide a same-data evidence chain in which point accuracy, marginal coverage, and joint event scores improve, yet queue benefits do not emerge in matched closed-loop control.
  \item We audit and correct a forecast--control interface, quantify the effective action space, and distinguish a successful synthetic positive control from negative results on real reconstructed demand.
  \item We report historical failures, paired-day uncertainty, metric trade-offs, and explicit stopping rules, demonstrating how a negative result can reveal reusable design constraints rather than being reduced to ``no improvement.''
\end{itemize}

\section{Related Work}

\subsection{Traffic forecasting and spatial structure}

Modern traffic forecasting models learn temporal dependence and graph structure through diffusion convolution, spatio-temporal graph convolution, adaptive adjacency, or attention \cite{li2018dcrnn,yu2018stgcn,wu2019graphwavenet,lim2021tft}. These models establish that traffic states are predictable and spatially coupled, but their common losses are usually averages over sensors and horizons. Such losses do not indicate whether the remaining errors are located at control-sensitive approaches or whether the prediction target distinguishes feasible actions. Our study therefore emphasizes two nested targets---entry-road totals and entry-to-next-road movements---rather than introducing a new deep forecasting architecture.

\subsection{Probabilistic forecasts and conformal calibration}

Probabilistic predictions should be evaluated by both sharpness and calibration using proper scoring rules \cite{gneiting2007proper}. Conformalized quantile regression (CQR) adjusts model-based quantile intervals using held-out residuals \cite{romano2019cqr}, while adaptive methods address sequential distribution shift \cite{gibbs2021adaptive,angelopoulos2023conformal}. Marginal coverage, however, is not a guarantee for a full spatial-temporal trajectory or for a high-demand condition. Dependence reconstruction methods such as ensemble copula coupling preserve marginal distributions while restoring cross-dimensional rank structure \cite{schefzik2013ecc}. We evaluate these layers separately and avoid interpreting one-dimensional coverage as joint operational safety.

\subsection{Signal control and forecast information}

MaxPressure control selects phases using upstream--downstream pressure and has strong network-level stability motivation \cite{varaiya2013maxpressure,wongpiromsarn2012distributed}. PressLight and CoLight integrate pressure and network cooperation into reinforcement learning \cite{wei2019presslight,wei2019colight}; CityFlow provides a platform for large-scale traffic-control experiments \cite{zhang2019cityflow}. Store-and-forward models offer computationally efficient network control and optimization \cite{aboudolas2009store}. Yet the value of a forecast depends on whether the internal dynamics, horizon, switching constraints, and action search can use future information. We operationalize this requirement through a matched oracle test rather than treating an oracle as an automatic upper bound.

\subsection{Prediction versus decision loss}

Decision-focused learning explicitly optimizes downstream outcomes rather than predictive error alone \cite{wilder2019decision,elmachtoub2022smart}. In traffic signal control, the mapping from error to decision loss is additionally discontinuous because actions are discrete, minimum-green and switching-loss constraints are active, and finite storage produces thresholds. Our contribution is not a new end-to-end decision-focused learner. Instead, we decompose the conversion path into testable interface conditions so that forecast failure and controller failure can be separated.

\section{Data and Experimental Protocol}

\subsection{Dataset and subnetwork}

The source data are the high-resolution Xuancheng traffic records released with refined road networks for hierarchical control \cite{ma2026city}. After file-level cleaning, we retain 29 valid days in April 2023. April 9 is excluded because it is byte-identical to April 8. Some routes require map matching, filtering, and shortest-path completion. Accordingly, the experiment is described as closed-loop simulation with demand reconstructed from real traffic records, not as an online field deployment.

The induced subnetwork contains nine signalized intersections, seven boundary entry roads, nineteen internal roads, and eight exit roads. Entry demand is aggregated in one-minute intervals. We additionally define twenty topology-valid movements by pairing each boundary entry road with its immediate next road. A movement timestamp is the vehicle's source-departure minute. It is a proxy for boundary arrival, not a direct boundary-arrival observation.

\subsection{Chronological split and leakage controls}

The date split was frozen before final evaluation (Table~\ref{tab:split}). Thirteen days are used for training, four for forecast/model validation, five for probability calibration, and seven for final testing. Due to availability of complete control assets, control-structure validation uses April 15 and April 18. The calibration partition is used for CQR and dependence templates but not for control parameter selection. Test dates are evaluated only after the relevant method is frozen.

\begin{table}[t]
\caption{Frozen chronological split.}
\label{tab:split}
\centering
\begin{tabular}{lll}
\toprule
Role & Dates in April 2023 & Number of days \\
\midrule
Training & 1--8 and 10--14 & 13 \\
Validation & 15--18 & 4 \\
Calibration & 19--23 & 5 \\
Test & 24--30 & 7 \\
\bottomrule
\end{tabular}
\end{table}

Each forecast uses a 60-minute history and predicts the next 15 one-minute steps. Features include lags 0, 1, 4, 14, 29, and 59 minutes; rolling statistics over 5, 15, 30, and 60 minutes; and calendar features. Crucially, the complete count of minute $m$ becomes observable only after minute $m$. The final controller therefore uses the forecast generated at origin $m-1$ for any decision in minute $m$. Arrivals already observed during minute $m$ are subtracted from its predicted total, and only the residual is distributed across unobserved five-second slots.

\subsection{Evaluation metrics}

Forecast metrics include MAE, RMSE, weighted absolute percentage error, bias, pinball loss, interval coverage, interval width, energy score, variogram score, and Brier scores for joint high-demand events. The main control metric is queue vehicle-seconds accumulated over monitored roads during the evaluation hour. Secondary metrics are exit throughput, completed-trip travel time, maximum road occupancy, and the fraction of monitored road-time with occupancy at or above 80\% of storage capacity.

For control comparisons, the primary effect is the ratio of seven-day arithmetic means. For a metric $M$ where lower is better,
\begin{equation}
\Delta_M=\frac{\overline{M}_{\mathrm{base}}-\overline{M}_{\mathrm{candidate}}}{\overline{M}_{\mathrm{base}}}\times 100\%.
\label{eq:effect}
\end{equation}
Thus positive values always indicate improvement. Uncertainty is summarized by 50,000 paired bootstrap resamples over dates with a fixed seed. We also report exact two-sided sign tests and leave-one-day-out ranges. With only seven independent test days, all inferential quantities are exploratory and no multiplicity correction is applied.

\section{Forecasting and Uncertainty Methods}

\subsection{Entry and movement point forecasts}

Entry-level forecasting compares persistence, a historical mean by weekday/minute, and ExtraTrees. Hyperparameters are selected by validation MAE, after which the model is refit on training plus validation dates for test evaluation. At the movement level, we compare movement historical means, entry forecasts allocated by global historical shares, entry forecasts allocated by rolling shares, direct ExtraTrees movement forecasts, and entry-reconciled direct forecasts.

The selected movement model is hierarchical rolling-share forecasting. It first predicts each of the seven entry totals and then allocates the total across topology-valid movements using the previous 60 minutes of movement composition. This construction preserves entry totals while adapting to short-term turning composition. The point metric is
\begin{equation}
\mathrm{MAE}=\frac{1}{N}\sum_{i=1}^{N}|y_i-\widehat{y}_i|.
\end{equation}

\subsection{Quantile forecasts and CQR}

LightGBM models estimate quantiles $\tau\in\{0.05,0.10,0.25,0.50,0.75,0.90,0.95\}$. Raw quantiles are sorted per target to remove crossings. For a central interval with miscoverage $\alpha$, CQR computes calibration scores
\begin{equation}
s_i=\max\left\{\widehat{q}_{\alpha/2}(x_i)-y_i,\ y_i-\widehat{q}_{1-\alpha/2}(x_i)\right\},
\end{equation}
and expands both endpoints by the empirical $(1-\alpha)$ quantile of $s_i$. Calibration is performed separately by entry, horizon, and interval level; the median remains unchanged.

We also diagnose an ex-post high-demand subset, defined by realized future demand exceeding the corresponding calibration-period 90th percentile. This subset is useful for locating tail errors but is not observable at decision time and is not claimed as a conditional-coverage guarantee.

\subsection{Dependence-aware scenarios}

For each origin, we generate 100 scenarios over $7\times 15=105$ entry--horizon targets. The correlated sampler draws the joint rank pattern from a single calibration residual block and maps it through the current CQR marginals. The independent sampler permutes the same marginal samples dimension by dimension. The maximum difference between sorted marginal samples is zero and the maximum difference in sample means is below $2.5\times10^{-5}$. Differences in joint scores can therefore be attributed primarily to dependence rather than marginal changes.

The joint events are (i) total 15-minute demand exceeding a calibration threshold, (ii) peak one-minute network demand exceeding its threshold, and (iii) persistent high demand across horizons. Brier scores are reported along with energy and variogram scores.

\begin{figure}[t]
  \centering
  \includegraphics[width=\textwidth]{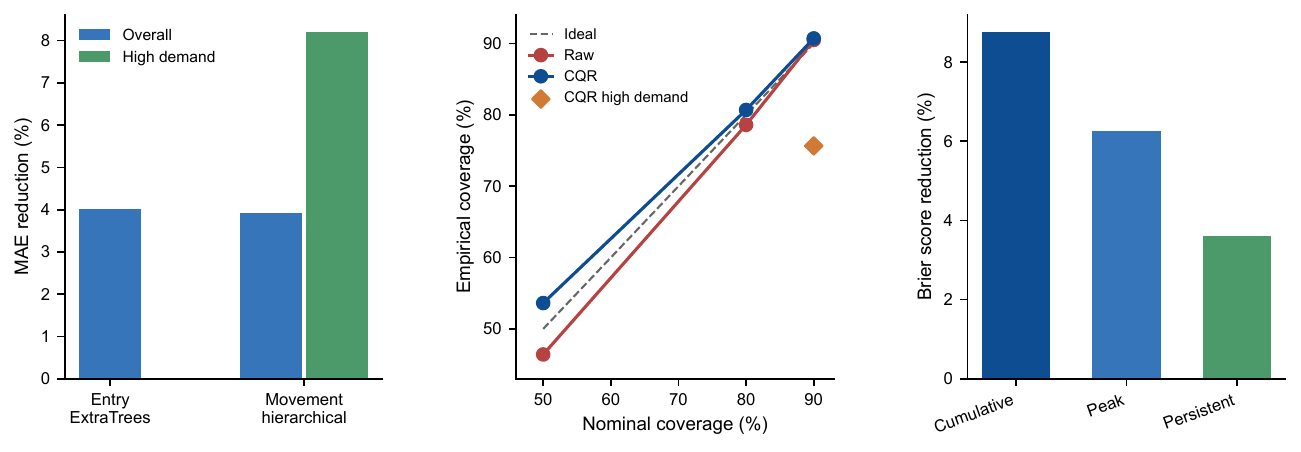}
  \caption{Forecast-layer results. Left: MAE reductions relative to historical-mean baselines; high-demand improvement is shown only for the selected movement model. Center: raw and CQR marginal coverage; the orange diamond is the ex-post high-demand subset at the 90\% interval. Right: Brier-score reductions from retaining cross-target dependence while keeping marginal samples fixed.}
  \label{fig:forecast}
\end{figure}

\section{Control Models and Interface Audit}

\subsection{Store-and-forward environment}

The closed-loop environment is an aggregate store-and-forward queue model \cite{aboudolas2009store}. Road storage is approximated from road length, lane count, and a fixed vehicle-space parameter. Saturation service is 0.5 vehicles per lane per second. The simulation begins at 06:30; 06:30--07:00 is warm-up and 07:00--08:00 is evaluation. Demand is injected every five seconds. Signal decisions are revisited every 30 seconds, with a 30-second minimum green and a five-second switching loss. Transfers are constrained by downstream storage.

The final rollout uses a 60-second horizon with 12 five-second steps. Its internal cost is
\begin{equation}
J(\mathbf{u})=\sum_{k=1}^{12}5\left[
\sum_{r\in\mathcal{R}_M}q_{r,k}+
\eta\sum_{r\in\mathcal{R}_M}C_r
\left(\max\left\{0,\frac{q_{r,k}}{C_r}-0.8\right\}\right)^2
\right],
\label{eq:rollout}
\end{equation}
where $q_{r,k}$ is road queue, $C_r$ is storage capacity, and $\eta=10$. We compare fixed-time control, MaxPressure, a rollout with no future arrivals, a rollout with causal forecasts, and a rollout with exact future source-departure events.

\subsection{Historical diagnostic chain}

The control study proceeded through deliberately retained versions. The first queue model produced an untrustworthy MaxPressure baseline. Adding five-second state advancement, minimum green, switching loss, and capacity-normalized pressure yielded a credible deterministic baseline: MaxPressure reduced queue vehicle-seconds by 20.39\% relative to fixed time. Yet adding entry-demand forecasts changed queue performance by $-0.15$\% relative to MaxPressure.

An entry-total finite-horizon interface selected a zero-minute horizon on validation, indicating that entry totals could not identify phase-specific future demand. Movement forecasts repaired this semantic mismatch, but an additive movement-pressure controller still selected zero forecast weight. Even when the additive controller received perfect future movement totals and independently selected its validation parameters, queue improvement was $-0.22$\% and only two of seven test days improved.

\begin{figure}[t]
  \centering
  \includegraphics[width=\textwidth]{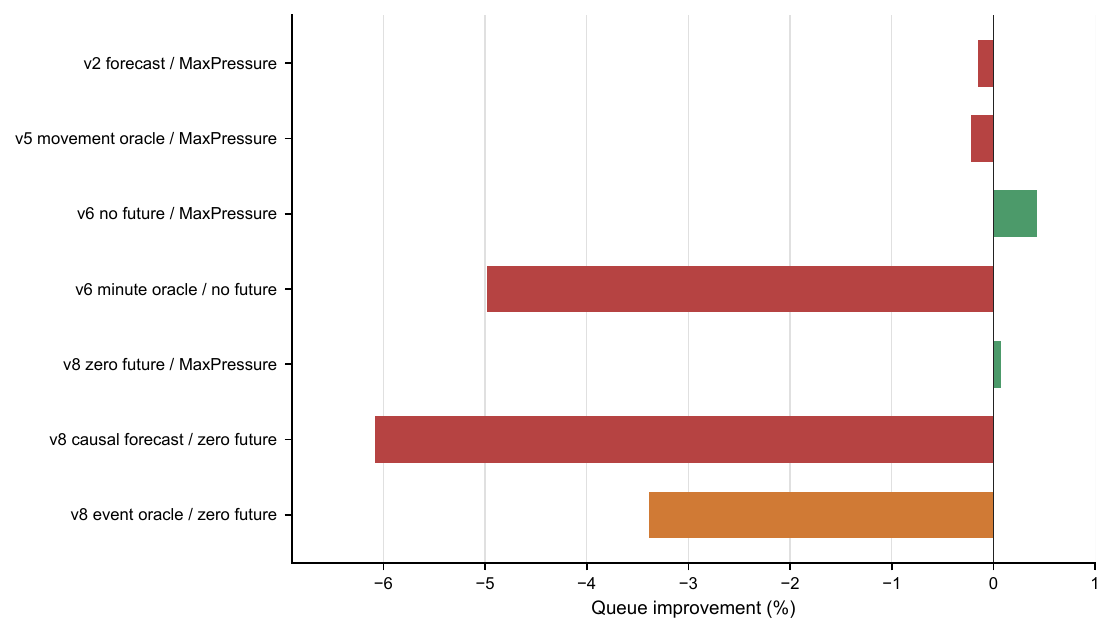}
  \caption{Queue effects across the diagnostic chain. Each bar uses the baseline written in its label; values are therefore not one common ablation scale. Historical v6 is shown as a diagnostic, while v8 is the frozen corrected evaluation.}
  \label{fig:evolution}
\end{figure}

A later network rollout explicitly advanced movement queues, storage, switching loss, and internal propagation. Its no-future version improved queue vehicle-seconds by 0.43\% relative to MaxPressure, while its minute-level oracle worsened queue vehicle-seconds by 4.98\% relative to the no-future rollout. These historical results motivated an interface audit and are not treated as a strict oracle bound (Fig.~\ref{fig:evolution}).

\subsection{Temporal audit and event oracle}

The audited simulation loop first inserts arrivals for the current five-second step, then observes the queue state, selects a phase, and executes service. The earlier oracle added current-minute demand again in the first projection slot, although the current-step arrivals were already in the observed state. It also called the minute-$m$ forecast whose lag-0 feature required the completed count of minute $m$.

The corrected event oracle sets projection slot 0 to zero new arrivals and inserts events from \texttt{current\_step}$+k$ in slot $k>0$. The causal forecast uses origin $m-1$. Within minute $m$, already observed movement counts are subtracted from the predicted minute total, and the nonnegative residual is divided only across the remaining unobserved seconds. This convention aligns the forecast with the simulator's add-before-service semantics.

\subsection{Effective action space and joint search}

Signal phases are deduplicated by their nonempty movement sets. Of nine nominally controlled intersections, seven have only one distinct nonempty action; two intersections have eight each. The effective one-stage joint action space therefore contains only $8\times 8=64$ combinations. The final controller enumerates all feasible joint actions at unlocked intersections rather than performing one coordinate-descent sweep. This audit reveals that nominal network size can substantially overstate decision freedom and forecast value.

\subsection{Positive control and dynamics consistency}

We construct a synthetic intersection with two competing movements, similar current queues, and a known burst arriving to the second movement. Without future arrivals, the optimal action serves movement 1; with exact future arrivals, exhaustive search and coordinate search both serve movement 2. Under the same future realization, the future-aware action reduces the internal rollout cost from 3195.8 to 1230.0, a 61.5\% reduction. This positive control establishes that the search and cost can respond to future information in a state where such information should matter.

We also compare one projected service step with a packet-level first-in-first-out (FIFO) step. The models match exactly when only one movement is eligible. When two movements compete for the same service budget, the rollout allocates proportionally while the executable simulator serves packets in FIFO order; the movement-queue $L_1$ difference is 2.5 vehicles in the focused test. The oracle is thus an information diagnostic within an approximate rollout, not a global control upper bound.

\section{Results}

\subsection{Forecast accuracy and reliability}

Table~\ref{tab:forecast_results} summarizes the main forecast-layer outcomes. Entry ExtraTrees reduces MAE from 1.9805 to 1.9007, a 4.03\% improvement over the historical mean. The hierarchical rolling-share movement model reduces MAE from 0.8446 to 0.8115, a 3.92\% improvement, and reduces high-demand MAE by 8.20\%. These gains show that source demand is predictable and that movement composition adds action-relevant spatial resolution.

\begin{table}[t]
\caption{Main forecast-layer results on the frozen test period. Positive improvement means lower error or Brier score.}
\label{tab:forecast_results}
\centering
\begin{tabular}{lrr}
\toprule
Result & Value & Improvement \\
\midrule
Entry ExtraTrees MAE & 1.9007 & 4.03\% \\
Movement hierarchical MAE & 0.8115 & 3.92\% \\
Movement high-demand MAE & --- & 8.20\% \\
CQR 90\% marginal coverage & 90.72\% & --- \\
CQR 90\% high-demand coverage & 75.66\% & --- \\
Correlated cumulative-event Brier & 0.04030 & 8.75\% \\
Correlated peak-event Brier & 0.04419 & 6.27\% \\
Correlated persistent-event Brier & 0.05102 & 3.61\% \\
\bottomrule
\end{tabular}
\end{table}

CQR reaches 80.69\% and 90.72\% empirical coverage for nominal 80\% and 90\% marginal intervals. However, 90\% coverage falls to 75.66\% on the ex-post high-demand subset, 15.06 percentage points below overall coverage. The finding is a tail diagnostic, not a conditional-coverage guarantee.

Dependence-aware scenarios improve joint event scores while leaving marginals essentially unchanged. Cumulative-event Brier score decreases from 0.04416 to 0.04030 (8.75\%); peak-event Brier score decreases from 0.04715 to 0.04419 (6.27\%); and persistent-event Brier score decreases from 0.05293 to 0.05102 (3.61\%). The energy and variogram scores also improve slightly. The forecast layer therefore contains genuine gains in accuracy, marginal reliability, and dependence-sensitive events.

\subsection{Corrected validation gate}

After the interface and action search are frozen, the corrected controller is first run on the two complete control-validation days. Event-oracle queue improvement relative to the no-future rollout is $-1.38$\%; causal-forecast improvement is $-2.56$\%. The oracle gate is therefore \texttt{NO\_GO}. We nevertheless perform one final test evaluation because the study objective is to quantify the frozen negative result, not to tune another controller. No subsequent control parameter is changed using test outcomes.

\subsection{Frozen closed-loop evaluation}

Table~\ref{tab:v8} reports the final seven-day means. MaxPressure strongly improves over fixed time. The no-future rollout is almost tied with MaxPressure on queue vehicle-seconds (0.08\% improvement) and increases throughput, but it has more spillback exposure and longer completed-trip travel time. The causal forecast has the highest queue among the adaptive controllers. The event oracle reduces spillback relative to the no-future rollout but does not reduce queue or improve throughput.

\begin{table}[t]
\caption{Frozen v8 test results averaged over seven dates.}
\label{tab:v8}
\centering
\small
\begin{tabular}{lrrrr}
\toprule
Controller & Queue veh.-s & Throughput & Spillback & Travel time (s) \\
\midrule
FixedTime & 1,069,814 & 2,967.9 & 13.63\% & 247.0 \\
MaxPressure & 851,685 & 3,072.8 & 9.59\% & 170.2 \\
RolloutZeroFuture & 851,022 & 3,102.7 & 10.37\% & 182.4 \\
RolloutCausalForecast & 902,821 & 3,028.2 & 10.33\% & 171.8 \\
RolloutEventOracle & 879,866 & 3,090.4 & 9.97\% & 183.2 \\
\bottomrule
\end{tabular}
\end{table}

Relative to RolloutZeroFuture, RolloutCausalForecast changes queue, throughput, and spillback by $-6.09$\%, $-2.40$\%, and $+0.31$\%, respectively. RolloutEventOracle changes the same metrics by $-3.39$\%, $-0.39$\%, and $+3.78$\%. Thus future information changes decisions and metric trade-offs, but it does not produce a joint queue-throughput gain under the frozen objective and dynamics.

\begin{figure}[t]
  \centering
  \includegraphics[width=0.86\textwidth]{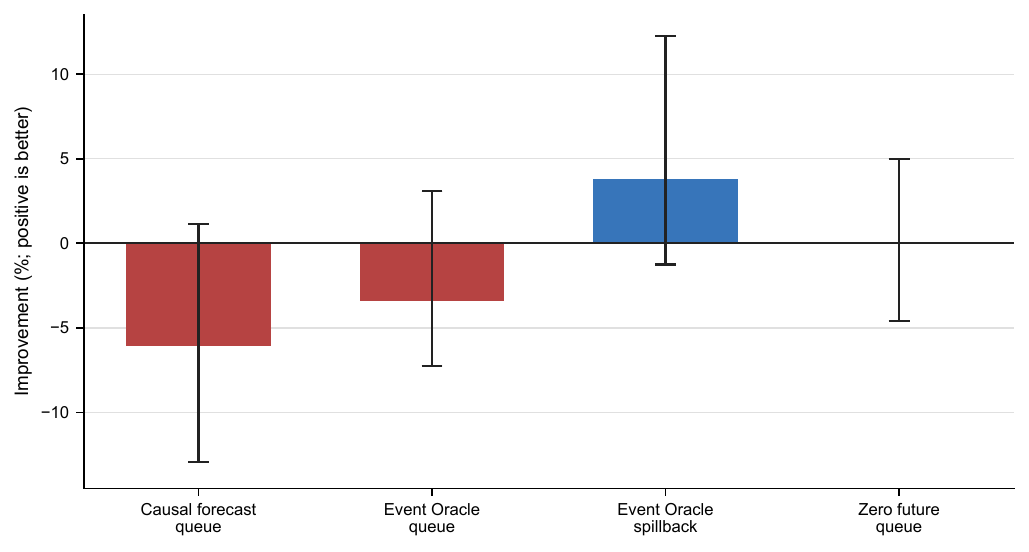}
  \caption{Frozen v8 effects. Positive values indicate improvement. Error bars are exploratory 95\% paired-day bootstrap intervals from 50,000 resamples of seven test dates.}
  \label{fig:v8}
\end{figure}

The causal-forecast queue effect is $-6.09$\% with a 95\% paired bootstrap interval of $[-12.94\%,1.15\%]$ and improvement on three of seven days. The event-oracle queue effect is $-3.39$\% with interval $[-7.24\%,3.08\%]$ and improvement on four days. Event-oracle spillback improvement is 3.78\% with interval $[-1.25\%,12.26\%]$ and improvement on five days. All intervals cross zero. The defensible statement is that we do not observe stable queue benefits; the data do not show that forecasts must harm control or that future information is generally worthless.

\section{Discussion}

\subsection{Four constraints on forecast-to-decision value}

\paragraph{Temporal observability.}
A controller may use only information available when it acts. The earlier interface used minute-$m$ lag-0 information before minute $m$ was complete and double-counted arrivals in the first rollout slot. Correcting these errors changes the diagnostic interpretation. Because negative queue results remain after correction, leakage was not the only limitation, but an unaudited oracle comparison would have been unreliable.

\paragraph{Action identifiability.}
Entry totals cannot distinguish phase-specific turning demand. Movement forecasts repair this semantic mismatch, yet only two of nine intersections have more than one effective action. Most nominally controlled nodes cannot respond differently to a better forecast. The potential decision value is therefore bounded jointly by information resolution and control authority.

\paragraph{Dynamics consistency.}
The synthetic positive control proves that the rollout can exploit future information in a deliberately sensitive state. In the reconstructed real-demand experiments, proportional movement service differs from FIFO execution when movements compete, and source-departure time differs from actual boundary-arrival time. A more informative input can make an optimizer act more aggressively on a biased internal model. This is a plausible mechanism, not a separately identified causal effect.

\paragraph{Objective alignment.}
The event oracle reduces spillback but increases queue vehicle-seconds. It is therefore inaccurate to say that the oracle has no effect. Rather, it shifts the efficiency--risk trade-off under a cost that combines queue and an 80\%-capacity spillback penalty. Queue is the frozen primary admission metric; switching to spillback after observing results would be outcome selection.

\subsection{Why forecast gains and control losses can coexist}

MAE measures average numerical proximity, CQR coverage measures a long-run marginal frequency, and Brier score measures probability quality for a specified event. Closed-loop value depends on state, action, constraints, dynamics, and objective. A 4\% MAE gain need not occur at the two intersections with effective phase choice or near a storage threshold. Likewise, improved dependence modeling matters only after the control structure passes an oracle-value gate. We therefore stop before running a full stochastic or CVaR controller: without positive oracle queue value, such an experiment would confound probabilistic quality with structural model error.

\subsection{Implications for traffic forecasting}

Offline prediction studies should supplement average errors with control-sensitive subgroup analyses, action-aligned targets, and a matched oracle test. High-demand reliability should ideally be grouped by decision-time observable states such as current queue, current load, predicted load, or action sensitivity. Our ex-post high-demand grouping locates tail errors but cannot support an online conditional guarantee.

\subsection{Implications for traffic control}

Before comparing forecast-aware and forecast-free controllers, researchers should establish a credible deterministic baseline, audit the time interface, and count the effective actions. An oracle is not automatically an upper bound; it is perfect information passed through a particular dynamics model, horizon, objective, and optimizer. When that pipeline is misspecified, an oracle can perform worse in execution. The correct label is ``future-information diagnostic within the current controller,'' not global optimum.

\subsection{Value of a reproducible negative result}

If only the final v8 result were reported, the source of failure would remain ambiguous. The retained chain shows that (i) a weak deterministic baseline must be repaired first; (ii) entry totals are not phase-identifying; (iii) movement resolution alone does not rescue an additive pressure interface; (iv) explicit rollout dynamics can improve no-future planning but still have a faulty time interface; and (v) after temporal correction, exact action search, and a successful positive control, stable queue value is still absent on real reconstructed demand. This sequence turns ``no gain'' into an auditable set of design boundaries.

\section{Limitations and Validity Threats}

The study covers one city and one month, with seven independent test dates. Neither bootstrap resampling nor additional simulation seeds create new real days. External validity across seasons, cities, and network structures remains open.

The demand is reconstructed from real records, but route completion, subnetwork extraction, and source-departure timing introduce measurement error. The event oracle knows future source departures, not true future boundary arrivals. The control environment is an aggregate queue model without microscopic car following, lane changing, lane-level queues, or peripheral background traffic. The findings are not CityFlow validation and are not evidence of field performance.

Only two intersections provide effective multi-action freedom. This fact is an important mechanism in the current case but restricts generalization to networks with richer control authority. The final rollout fixes one action over the 60-second horizon and uses proportional service internally. A multi-stage FIFO-consistent optimizer may produce different results. However, the current test set has now been consumed; such a controller must be preregistered and evaluated using a new split or external data rather than tuned on these test outcomes.

Finally, we do not evaluate a full dependence-aware stochastic MPC after the oracle gate fails. Consequently, the study does not claim that every risk-aware controller is ineffective. It shows that the available probabilistic improvements have no validated structural path to queue benefit in the audited controller.

\section{Conclusion}

This paper studies a routinely assumed but rarely audited proposition: better traffic forecasts should lead to better traffic decisions. On reconstructed multi-day demand from Xuancheng, entry and movement point forecasts improve, marginal conformal intervals approach nominal coverage, and dependence-aware scenarios improve joint high-demand event scores. None of these successes automatically becomes a stable queue benefit.

The interface audit sharpens the negative result. We remove decision-time leakage, construct a five-second event oracle, enumerate the 64 effective joint actions, and verify a synthetic state where future information reduces rollout cost by 61.5\%. Yet on frozen real-demand test dates, the causal forecast and event oracle do not improve queue vehicle-seconds relative to the matched no-future rollout; the oracle instead reveals a queue--spillback trade-off.

Forecast-to-decision value is therefore not a property of accuracy alone. It is jointly constrained by temporal observability, action identifiability, dynamics consistency, and objective alignment. The layered admission protocol offers a conservative workflow: validate forecasts, audit availability and action semantics, verify a positive control, require matched oracle value, and only then evaluate frozen forecasts or expensive risk-aware controllers. Future work should use directly estimated boundary arrivals, networks with richer effective action spaces, and multi-stage FIFO-consistent control on new evaluation data.

\section*{Reproducibility Statement}

All data-processing, forecasting, calibration, scenario-generation, control, and paired-day analysis stages use fixed chronological splits and seeds. The final interface and control implementation pass eleven focused unit tests; the interface audit passes eight independent checks and the frozen v8 experiment passes thirteen independent checks. Raw data are not modified. The calibration partition is not used for control selection, and no v8 control parameter is adjusted after test evaluation.

Core code, focused tests, and compact result summaries: \url{https://github.com/ljnvs/traffic}. The public repository does not redistribute source traffic records; the source dataset is described in \cite{ma2026city}.

\bibliographystyle{unsrtnat}
\bibliography{references}

\end{document}